\pdfoutput=1

\documentclass[11pt]{article}

\usepackage[final]{emnlp2023}

\usepackage{times}
\usepackage{latexsym}
\usepackage{amsmath}
\usepackage{amsfonts}
\usepackage{bbm}
\usepackage{multirow}
\usepackage{booktabs}
\usepackage{graphicx}
\usepackage{inconsolata}
\usepackage{tabu}
\usepackage{makecell}

\usepackage[T1]{fontenc}

\usepackage[utf8]{inputenc}

\usepackage{microtype}

\title{On the Relation between Sensitivity and Accuracy in In-Context Learning}

\author{Yanda Chen\textsuperscript{1}~~~~~~Chen Zhao\textsuperscript{2}~~~~~~Zhou Yu\textsuperscript{1}~~~~~~Kathleen McKeown\textsuperscript{1}~~~~~~He He\textsuperscript{2}\\
\textsuperscript{1}Columbia University,~~\textsuperscript{2}New York University\\\\
{\tt \{yanda.chen, kathy\}@cs.columbia.edu, cz1285@nyu.edu}\\
{\tt zy2461@columbia.edu, hehe@cs.nyu.edu}}

\begin{document}
\maketitle
\begin{abstract}
In-context learning (ICL) suffers from oversensitivity to the prompt, making it unreliable in real-world scenarios. We study the sensitivity of ICL with respect to multiple perturbation types. First, we find that label bias obscures the true sensitivity, and therefore prior work may have significantly underestimated ICL sensitivity. Second, we observe a strong negative correlation between ICL sensitivity and accuracy: predictions sensitive to perturbations are less likely to be correct. Motivated by these findings, we propose \textsc{SenSel}, a few-shot selective prediction method that abstains from sensitive predictions. Experiments on ten classification datasets
show that \textsc{SenSel} consistently outperforms two commonly used confidence-based and entropy-based baselines
on abstention decisions.
\end{abstract}



\section{Introduction}



\begin{figure*}[t]
\centering
\includegraphics[scale=0.203]{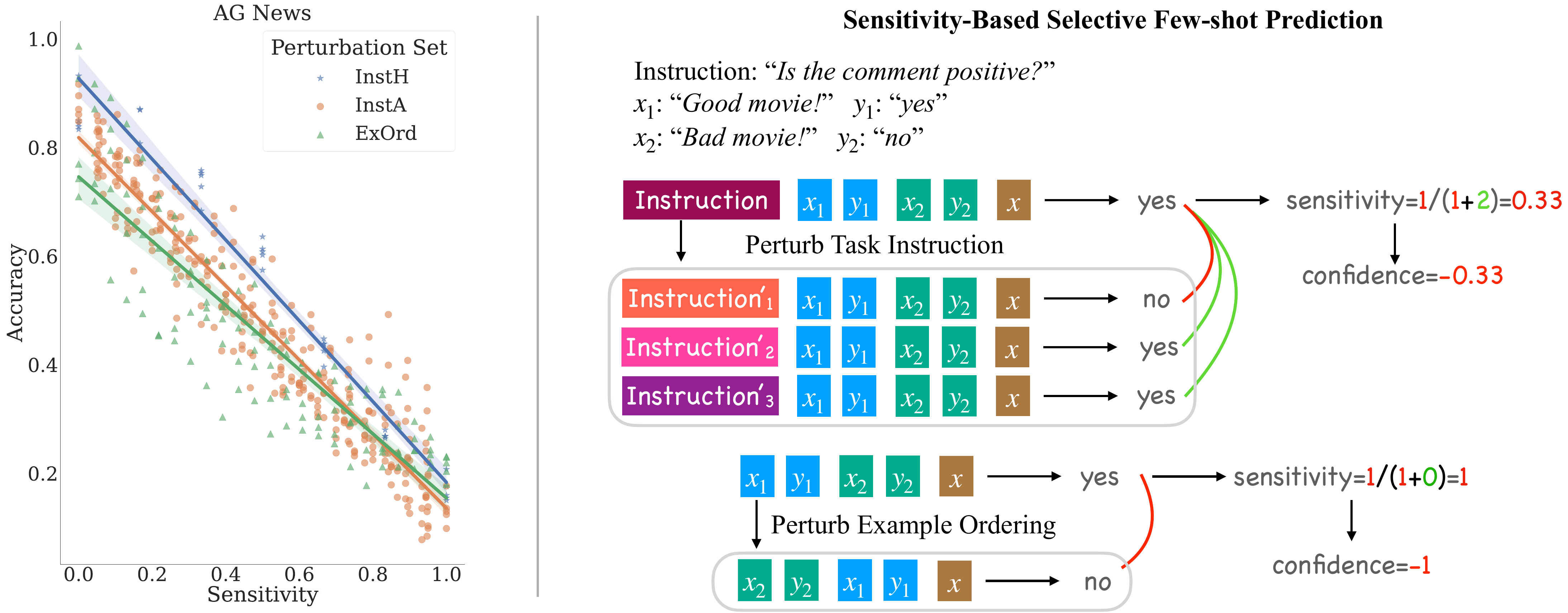}
\caption{\label{fig1} \textbf{ICL sensitivity-accuracy correlation} 
(left): We plot the prediction sensitivity against the prediction accuracy averaged over examples with that sensitivity. Different colors represent different perturbation sets (Section~\ref{sec:icl-sensitivity}), and color bands represent 95\% confidence intervals. We observe a significant negative correlation between the prediction sensitivity and accuracy of ICL. 
\textbf{\textsc{SenSel}} (right): \textsc{SenSel} measures the sensitivity of model predictions to prompt perturbations, and abstains from making predictions on examples with high sensitivity.}
\end{figure*}


Few-shot learning (FSL) refers to a system's ability to quickly learn a new task based on a few labeled examples.
Recently, in-context learning (ICL) has made significant progress in FSL, where a language model (LM) is prompted with a few demonstrated examples that enable it to make predictions for new examples without any gradient update.
However, a known issue of ICL is that it is oversensitive to the prompt \cite{zhao2021calibrate, perez2021true},
making it less reliable in practice.
%
Despite a near-universal acknowledgment of this issue, 
when and how the prediction is sensitive remains unclear \cite{min2022rethinking, kim2022ground}. 
This paper fills these gaps.

%
We conduct a systematic study of the ICL sensitivity to prompt perturbations. 
Specifically, we perturb the task instruction (by paraphrasing and noise injection)
and the in-context example orders.
We then measure the 
prediction sensitivity by the magnitude of model output changes due to the prompt perturbation.

Our first observation is that
the extent of sensitivity is significantly underestimated
due to \emph{label bias} in ICL:
LMs tend to assign a higher probability to a specific label regardless of the prompt \cite{zhao2021calibrate}
, thus appearing to make stable predictions.
Our study shows that the \textit{adjusted sensitivity} after mitigating label bias is up to $\mathbf{2.8x}$ of the \textit{raw sensitivity}.

%
After mitigating label bias,
we observe a negative correlation between the adjusted sensitivity and the accuracy of ICL: if a prediction is sensitive to prompt perturbations, then it is likely to be incorrect (Figure~\ref{fig1}~left). This finding aligns with our intuition that 
if the prediction is sensitive to the prompt that elicits the LM concept (e.g., sentiment) \cite{xie2022an},
then the example is likely not typical for that concept, and is thus more challenging.
Our experiments show a significant negative correlation of up to $\mathbf{-0.40}$ (Pearson) 
between ICL sensitivity and accuracy.


Given the above findings, a natural idea is to use sensitivity as a signal to 
abstain from making predictions on error-prone examples---an important mechanism to increase user trust when deploying ICL models to high-stakes domains such as healthcare \cite{Korngiebel2021ConsideringTP, Sezgin2022OperationalizingAI} and legal systems \cite{eliot2021generative}.
Our proposed method, Sensitivity-based Selective prediction (\textsc{SenSel}), uses sensitivity to make abstention decisions: the LM abstains on examples where its prediction is sensitive to prompt perturbations (Figure~\ref{fig1}~right). 
Compared to the common approach of training a separate model to make abstention decisions \cite{platt, geifman2019selectivenet, kamath-etal-2020-selective}, our approach does not require large amounts of labeled data and thus is more suitable for the few-shot setting. 

Our experiments show that sensitivity is a stronger signal than output probabilities for abstention.
\textsc{SenSel} consistently outperforms two baselines based on model probabilities (\textsc{MaxProb} and \textsc{Entropy}) by up to $\mathbf{+4.1}$ AUC points. 
Further analysis shows that \textsc{SenSel} and \textsc{MaxProb} are \emph{complementary}---\textsc{MaxProb} falters on high-sensitivity tasks because it relies on oversensitive model probabilities for abstention,
while \textsc{SenSel} capitalizes ICL sensitivity for abstention and hence works better on high-sensitivity tasks.\footnote{We released our code at \url{https://github.com/yandachen/ICLSensitivity}.}

Our contributions are as follows. \emph{(i)} We find that prior work may have significantly underestimated true ICL sensitivity if label bias is not mitigated beforehand. \emph{(ii)} We observe a strong negative correlation between ICL sensitivity and accuracy. \emph{(iii)} We propose a sensitivity-based selective prediction method \textsc{SenSel} that consistently outperforms two commonly used baselines based on model probabilities. 


\section{ICL Sensitivity Study}
\label{sec:sensitivity-study}

\begin{figure*}[t]
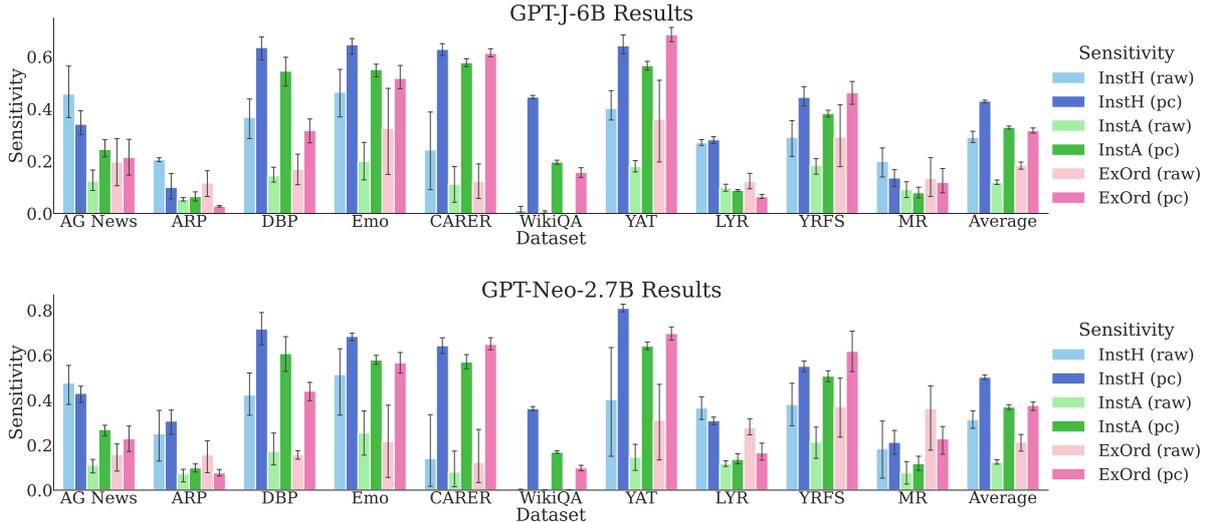

\centering
\includegraphics[scale=0.035]{figures/sensitivity_raw_pc_gpt-j-6B.pdf}\\
\vspace{1em}
\includegraphics[scale=0.035]{figures/sensitivity_raw_pc_gpt-neo-2.7B.pdf}
\caption{\label{fig:sensitivity} We compare the raw sensitivity with the adjusted sensitivity (label bias mitigated with PC). We observe that the adjusted sensitivity is consistently higher than the raw sensitivity for all three perturbation sets for both \textsc{GPT-J} and \textsc{GPT-Neo}. Error bars represent 95\% confidence intervals.}
\end{figure*}

\begin{table*}[t]
\fontsize{8}{8}\selectfont
\setlength{\tabcolsep}{3pt}
\centering
\begin{tabu}{c|c|cccccccccc|c}
\toprule
Model & Perturb Set & AG News & ARP & DBP & Emo & CARER & WikiQA & YAT & LYR & YRFS & MR & Avg \\
\midrule
\rowfont{\fontsize{9}{9}\selectfont}
\multirow{6}{*}{\textsc{GPT-J-6B}} & \multirow{2}{*}{\textsc{InstH}} & $-0.49$ & $-0.55$ & $-0.55$ & $-0.28$ & $-0.31$ & $0.04$ & $-0.35$ & $-0.61$ & $-0.27$ & $-0.49$ & $-0.39$ \\
\rowfont{\fontsize{7}{7}\selectfont}
&& $(0.04)$ & $(0.02)$ & $(0.10)$ & $(0.11)$ & $(0.01)$ & $(0.10)$ & $(0.02)$ & $(0.09)$ & $(0.04)$ & $(0.02)$ & $(0.02)$ \\
\cmidrule{2-13}
\rowfont{\fontsize{9}{9}\selectfont}
&\multirow{2}{*}{\textsc{InstA}} & $-0.40$ & $-0.39$ & $-0.65$ & $-0.27$ & $-0.31$ & $-0.18$ & $-0.55$ & $-0.41$ & $-0.25$ & $-0.39$ & $-0.38$ \\
\rowfont{\fontsize{7}{7}\selectfont}
&& $(0.02)$ & $(0.03)$ & $(0.08)$ & $(0.12)$ & $(0.04)$ & $(0.01)$ & $(0.01)$ & $(0.05)$ & $(0.03)$ & $(0.03)$ & $(0.01)$ \\
\cmidrule{2-13}
\rowfont{\fontsize{9}{9}\selectfont}
&\multirow{2}{*}{\textsc{ExOrd}} & $-0.38$ & $-0.46$ & $-0.82$ & $-0.17$ & $-0.32$ & $-0.09$ & $-0.51$ & $-0.52$ & $-0.26$ & $-0.47$ & $-0.40$ \\
\rowfont{\fontsize{7}{7}\selectfont}
&& $(0.08)$ & $(0.02)$ & $(0.02)$ & $(0.06)$ & $(0.06)$ & $(0.05)$ & $(0.07)$ & $(0.03)$ & $(0.04)$ & $(0.07)$ & $(0.03)$ \\
\midrule
\rowfont{\fontsize{9}{9}\selectfont}
\multirow{6}{*}{\textsc{GPT-Neo-2.7B}} & \multirow{2}{*}{\textsc{InstH}} & $-0.49$ & $-0.57$ & $-0.53$ & $-0.09$ & $-0.36$ & $-0.36$ & $-0.25$ & $-0.54$ & $-0.21$ & $-0.48$ & $-0.39$\\
\rowfont{\fontsize{7}{7}\selectfont}
&& $(0.04)$ & $(0.04)$ & $(0.14)$ & $(0.12)$ & $(0.04)$ & $(0.03)$ & $(0.02)$ & $(0.09)$ & $(0.07)$ & $(0.03)$ & $(0.02)$\\
\cmidrule{2-13}
\rowfont{\fontsize{9}{9}\selectfont}
&\multirow{2}{*}{\textsc{InstA}} & $-0.39$ & $-0.22$ & $-0.61$ & $-0.09$ & $-0.36$ & $-0.10$ & $-0.41$ & $-0.19$ & $-0.17$ & $-0.28$ & $-0.28$\\
\rowfont{\fontsize{7}{7}\selectfont}
&& $(0.03)$ & $(0.02)$ & $(0.13)$ & $(0.06)$ & $(0.05)$ & $(0.03)$ & $(0.03)$ & $(0.07)$ & $(0.07)$ & $(0.05)$ & $(0.02)$\\
\cmidrule{2-13}
\rowfont{\fontsize{9}{9}\selectfont}
&\multirow{2}{*}{\textsc{ExOrd}} & $-0.27$ & $-0.48$ & $-0.76$ & $-0.21$ & $-0.29$ & $-0.33$ & $-0.29$ & $-0.46$ & $-0.14$ & $-0.28$ & $-0.35$\\
\rowfont{\fontsize{7}{7}\selectfont}
&& $(0.06)$ & $(0.06)$ & $(0.04)$ & $(0.12)$ & $(0.06)$ & $(0.04)$ & $(0.14)$ & $(0.08)$ & $(0.04)$ & $(0.07)$ & $(0.04)$\\
\bottomrule
\end{tabu}
\caption{\label{tab:sensitivity-corr} We report the Pearson correlation coefficient (and its standard deviation in parenthesis) between ICL sensitivity and accuracy across five randomly sampled sets
of few-shot examples (label bias mitigated with PC). 
We observe a strong negative correlation between ICL sensitivity and accuracy for all perturbation sets and both models.  
}
\end{table*}

\begin{figure*}[t]
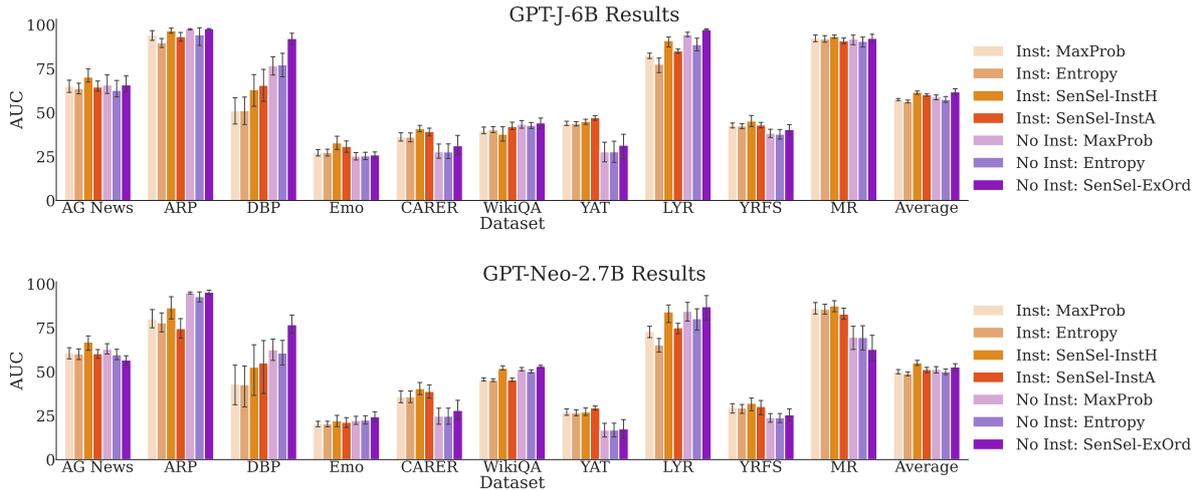

\centering
\includegraphics[scale=0.0325]{figures/sensel_auc_gpt-j-6B-pc.pdf}\\
\vspace{1em}
\includegraphics[scale=0.0325]{figures/sensel_auc_gpt-neo-2.7B-pc.pdf}
\caption{\label{fig:auc} We compare our \textsc{SenSel} method (label bias mitigated with PC) to the \textsc{MaxProb} baseline on abstention, measured by AUC score. \textsc{SenSel} consistently outperforms \textsc{MaxProb} on both the \textsc{Inst} and \textsc{No Inst} setting. 
}
\end{figure*}


In this section, we study the interplay between label bias and prediction sensitivity in ICL,
as well as the relation between sensitivity and accuracy.

\subsection{ICL Sensitivity}
\label{sec:icl-sensitivity}

\paragraph{Background}
In-context learning is a FSL method using LMs.
Given a test example $x$, we concatenate the task instruction $I$, a few ($K$) labeled examples $S = [(x_{\sigma(i)}, y_{\sigma(i)})]_{i=1}^K$ in $\sigma$ order,
and the test input $x$. 
The probability of each label is then assigned by the next-word probabilities from the LM.
We use $p_{\mathrm{LM}}(y \mid x, I, S, \sigma)$ to denote the prediction probabilities, and $f(x, I, S, \sigma) = \arg \max_y {p_{\mathrm{LM}}(y \mid x, I, S, \sigma)}$ to denote the predicted (most likely) label.

Despite its success,  ICL is known to be highly sensitive. 
Several methods have been proposed to address this issue.
\citet{zhou2022prompt} 
fine-tune LM to produce consistent predictions on various prompts, while \citet{chen-etal-2022-meta} and \citet{min-etal-2022-metaicl} meta-train models to perform ICL to reduce sensitivity. \citet{lu-etal-2022-fantastically} search for high-performance prompts that lead to less sensitive predictions. 
In contrast to these works, we connect ICL sensitivity to label bias and prediction accuracy,
and propose a new few-shot selective prediction approach based on sensitivity.

\paragraph{Measuring Sensitivity}
We measure prediction sensitivity by the magnitude of the changes in the predicted label when the prompt is perturbed.
We perturb the task instruction and the order of the in-context examples respectively.
Formally, we measure the sensitivity of a prediction $f(x, I, S, \sigma)$ with respect to perturbation set $P$ as
$$\frac{1}{|P|}\sum_{(I', S', \sigma') \in P}
    \mathbf{1}[f(x, I, S, \sigma) \neq f(x, I', S', \sigma')].$$

We use three perturbation sets.
\emph{Human Instruction Perturbation} (\textsc{InstH}) replaces the instruction with other human-written instructions for the same task;
\emph{Automatic Instruction Perturbation} (\textsc{InstA})  perturbs the task instruction by dropping out words and paraphrasing (details in Appendix~\ref{sec:appendix-setup-study});
%
\emph{Example Ordering Perturbation} (\textsc{ExOrd})
permutes
the ordering of the in-context examples.

\paragraph{Confounding with Label Bias}
One known issue 
of ICL is label bias, 
where LMs
assign a higher probability to a 
specific label regardless of the prompt, and hence 
appear to make stable predictions when the prompt is perturbed. 
Prior work mitigates label bias by adjusting the decision boundary. 
%
For example,
contextual calibration (CC) re-normalizes the predicted label distribution such that it is uniform given null examples \cite{zhao2021calibrate}.
%
%
%
Prototypical calibration (PC) 
clusters the LM's predictions,
maps each cluster to a label, and make predictions for new examples by their most likely cluster assignments \cite{han2022prototypical}.
\subsection{Experimental Setup}
We first compare the raw sensitivity with the adjusted sensitivity. 
We then compute the Pearson correlation coefficient \cite{freedman2007statistics} between the adjusted sensitivity and accuracy.

We run experiments on ten classification datasets covering sentiment classification, emotion classification, topic classification, and question-answering. See Appendix~\ref{sec:datasets} for dataset details.
We use \mbox{\textsc{GPT-J-6B}} \cite{gpt-j} and \mbox{\textsc{GPT-Neo-2.7B}} \cite{gpt-neo, gao2020pile} as our models. 
We describe additional implementation details in Appendix~\ref{sec:appendix-setup-study}.
For label bias mitigation, because the same observations hold for PC and CC, we report PC results in the main paper and CC results in Appendix~\ref{sec:additional-study-results}.




\subsection{Findings}
\label{sec:sensitivity-study-results}


\paragraph{Sensitivity is underestimated due to label bias.}
%
We report raw and adjusted sensitivity with respect to each perturbation set in Figure~\ref{fig:sensitivity}. We observe on both models and all three perturbation sets that ICL becomes more sensitive when
label bias is mitigated.
After prototypical calibration, the adjusted sensitivity is $\mathbf{99.0\%}$ higher.
Therefore, we argue that the true sensitivity may have been significantly underestimated if label bias is not mitigated.

Among the three perturbation sets, ICL is most sensitive to human instruction perturbations: the perturbations cause the predicted label to change 43\% of the time on \textsc{GPT-J-6B} and 50\% of the time on \textsc{GPT-Neo-2.7B} (after mitigating label bias).
This may be caused by the semantic difference in various human instructions for the same task, such as changing ``\textit{Is this product review positive?}'' to ``\textit{Based on this review, would the user recommend this product?}''.

\paragraph{Sensitivity is negatively correlated to accuracy.}
%
After mitigating label bias, we measure the Pearson correlation coefficient between sensitivity and accuracy (Table~\ref{tab:sensitivity-corr}).
We observe a significant negative correlation between sensitivity (with respect to all perturbation sets) and accuracy across datasets. The correlation is strongest for human instruction perturbations ($\mathbf{-0.39}$ on both models). 

\section{Sensitivity-based Selective Few-shot Prediction}
\label{sec:sensitivity-selective-prediction-method}


Motivated by the correlation between the sensitivity and accuracy of ICL, we propose \textsc{SenSel}---a selective few-shot prediction method based on sensitivity.

\paragraph{Problem Statement}
\label{sec:abstention-setup}
The goal of selective prediction is to \textit{abstain} on examples that the model is not confident about, to avoid presenting wrong predictions to users \cite{5222035, foundations2010}. 
%
Selective prediction methods score model confidence $C$ on each example, and abstain on examples with low prediction confidence ($C < \gamma$), where $\gamma$ is a threshold to control the trade-off between accuracy and coverage.

\paragraph{Sensitivity-based Selective Prediction}
\textsc{SenSel} scores ICL prediction confidence as the negative value of the prediction's sensitivity to prompt perturbations, and then abstains on 
examples whose confidence scores (i.e., negative sensitivity scores) are below a certain threshold $\gamma$.

\paragraph{Experiment Setup}
\label{sec:selective-eval}
For \textsc{SenSel}, we always use the adjusted sensitivity computed after mitigating the label bias. As writing good task instructions can be hard \cite{gao-etal-2021-making}, 
we experiment with two settings: \textsc{Inst} (a task instruction is available), and \textsc{No Inst} (no task instruction is available). We perturb the task instruction in the \textsc{Inst} setting (\textsc{SenSel-InstH}, \textsc{SenSel-InstA}), and perturb the example ordering in the \textsc{No Inst} setting (\textsc{SenSel-ExOrd}).
We compare  \textsc{SenSel} to two simple yet strong baselines, \textsc{MaxProb}, which uses the maximum output probability over the labels as the confidence score \cite{hendrycks2017a, lakshminarayanan2017simple}, and \textsc{Entropy}, which uses the negative value of the entropy of the output probabilities over the labels as the confidence score \cite{80269}.
%
%
We evaluate the effectiveness of selective prediction methods with the area under the F1-Coverage curve (AUC), which measures the average F1-score at different coverage rates \cite{kamath-etal-2020-selective}.
For label bias mitigation, since the same conclusion holds for PC and CC, we report the results for PC in the main paper and the results for CC in Appendix~\ref{sec:additional-sensel-results}. 
\paragraph{Results}

According to Figure~\ref{fig:auc},
 \textsc{SenSel} consistently outperforms \textsc{MaxProb} and \textsc{Entropy}. Among the three perturbation sets, \textsc{SenSel} with human-written instruction perturbations performs the best (outperforming \textsc{MaxProb} by an average margin of $\mathbf{+4.1}$ AUC points on \textsc{GPT-J-6B} and $\mathbf{+5.0}$ AUC points on \textsc{GPT-Neo-2.7B}), which is consistent with our sensitivity study that sensitivity to human-written instructions has the strongest correlation with accuracy. Even when instructions are not available, \textsc{SenSel-ExOrd} outperforms \textsc{MaxProb} and \textsc{Entropy} consistently by an average margin of $\mathbf{+3.0}$ AUC points on \textsc{GPT-J-6B} and $\mathbf{+1.4}$ AUC points on \textsc{GPT-Neo-2.7B}.
To understand how well \textsc{SenSel} and \textsc{MaxProb} perform on different tasks, we analyze the two methods on tasks with different prediction sensitivity.
Specifically, we measure the correlation between task sensitivity and task abstention performance (measured by the AUC of each abstention method minus that of a random abstention baseline).
Results show that \textsc{MaxProb} works better on tasks with low prediction sensitivity (Pearson correlation $-0.17$), while \textsc{SenSel} works better on tasks with high prediction sensitivity (correlation $+0.28$) (Figure~\ref{fig:sensitivity}, Figure~\ref{fig:auc}).
Hence, \textsc{SenSel} and \textsc{MaxProb} are \emph{complementary}: \textsc{MaxProb} falters on high-sensitivity tasks (e.g., DBP) because it relies on oversensitive model probabilities for abstention,
while \textsc{SenSel} capitalizes ICL sensitivity for abstention and hence works even better on high-sensitivity tasks. 

\section{Conclusion}
While ICL sensitivity is a widely-known issue, its relation to other variables is not studied.
This work first conducts a comprehensive study, and finds that ICL sensitivity is negatively correlated with accuracy when label bias is mitigated. Based on this observation, we develop a few-shot selective prediction method that abstains on highly sensitive predictions.
Our results show that ICL sensitivity exhibits a useful pattern---it reflects how confidently an LM understands the task.

There are many open questions for future work.
First, our study of the sensitivity-accuracy relation is \emph{correlational} but not \emph{causal}. Future work should explore causal experiments to study whether ICL predictions are sensitive because they are uncertain.
Second, it remains unclear \textit{why} sensitivity is negatively correlated with accuracy in ICL, which requires a better understanding of the mechanism of ICL. 
Third, our work mainly focuses on the text \emph{classification} tasks. Future work can further explore other tasks such as text generation and question answering with structured output.

\section{Limitations}
First, our study of the sensitivity-accuracy relation is correlational but not causal. Future work should explore causal experiments to study whether ICL predictions are sensitive because they are incorrect. Second, it remains unclear why sensitivity is negatively correlated with accuracy in ICL, which requires a better understanding of the mechanism of ICL. Third, our work mainly focuses on the text classification tasks. Future work can further explore other tasks such as text generation.
\section{Acknowledgements}
This research is supported in part by the Office of the Director of National Intelligence (ODNI), Intelligence Advanced Research Projects Activity (IARPA), via the HIATUS Program contract \#2022-22072200005. The U.S. Government is authorized to reproduce and distribute reprints for governmental purposes notwithstanding any copyright annotation therein. This work was funded in part by the US Department of Defense under the DARPA CCU program. Any opinions expressed herein are those of the authors and do not necessarily reflect the views of the U.S. Department of Defense, ODNI, IARPA, or any other agency of the U.S. Government.
This research is supported in part by AWS AI, Samsung Advanced Institute of Technology (under the project Next Generation Deep Learning: From Pattern Recognition to AI), and Open Philanthropy.
YC is supported by an Avanessians Doctoral Fellowship. CZ is supported by Shanghai Frontiers Science Center of Artificial Intelligence and Deep Learning, NYU Shanghai.
\newpage

\bibliography{custom}
\bibliographystyle{acl_natbib}

\clearpage

\appendix
\section{Datasets}
\label{sec:datasets}
We study ICL sensitivity and few-shot selective prediction on the following datasets: AG News \cite{Zhang2015CharacterlevelCN}, Amazon Review Polarity (ARP, \citet{10.1145/2507157.2507163}), DBPedia14 (DBP, \citet{dbpedia}), Emo2019 (Emo, \citet{chatterjee-etal-2019-semeval}), Contextualized Affect Representations for Emotion Recognition (CARER, \citet{saravia-etal-2018-carer}), Wiki Question Answering (WikiQA, \cite{yang-etal-2015-wikiqa}), Yahoo Answers Topics (YAT, \citet{zhang2015text}), Large Yelp Review (LYR, \citet{Zhang2015CharacterlevelCN}), Yelp Reviews Full Star (YRFS, \citet{zhang2015text}), and Rotten Tomatoes Movie Review (MR, \citet{pang-lee-2005-seeing}).

\section{Sensitivity Study Implementation Details}
\label{sec:appendix-setup-study}
\paragraph{ICL} We set the number of shots $K$ to four because the performance flattens out beyond four examples in our setting. All results are averaged over five randomly sampled sets of few-shot examples.

\paragraph{Label Bias} To reduce label bias, for CC we follow \citet{zhao2021calibrate} and use the empty string, the ``[MASK]'' token, and the ``N/A'' string as the null examples. For PC, similar to \citet{han2022prototypical} we use 1000 unlabeled examples for clustering.

\paragraph{Perturbation Set}
For human instruction perturbation, we use task instructions from PromptSource \cite{bach-etal-2022-promptsource}, which provides on average 7 task instructions for each task. For automatic instruction perturbation,  we generate 10 perturbed instructions by randomly dropping out 20\% of the tokens in the instruction, and another 10 perturbed instructions by using a neural paraphrase model. We use a T5 model fine-tuned on the Google PAWS dataset \cite{zhang-etal-2019-paws} as the paraphrase model and decode with nucleus sampling of top-$p=0.9$.

\section{Additional Results}
\subsection{ICL Sensitivity Study}
\label{sec:additional-study-results}

\begin{figure*}[t]
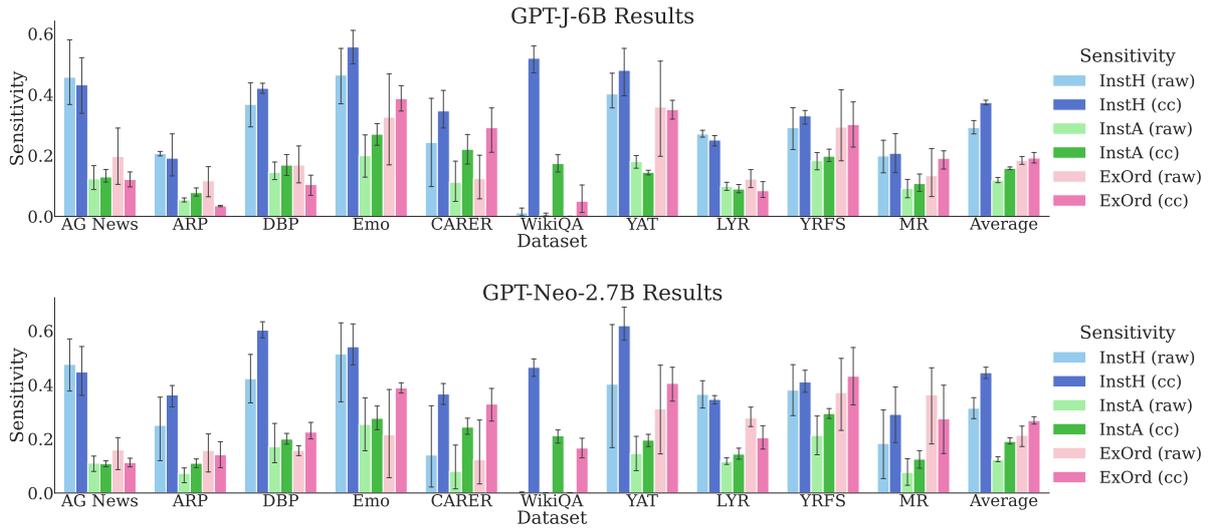

\centering
\includegraphics[scale=0.035]{figures/sensitivity_raw_cc_gpt-j-6B.pdf} \\
\vspace{1em}
\includegraphics[scale=0.035]{figures/sensitivity_raw_cc_gpt-neo-2.7B.pdf}
\caption{\label{fig:sensitivity_cc1} We compare the raw sensitivity with the adjusted sensitivity (label bias mitigated with CC). We observe that the adjusted sensitivity is consistently higher than the raw sensitivity for all three perturbation sets (\textsc{InstH}: Human Instruction Perturbation, \textsc{InstA}: Automatic Instruction Perturbation, and \textsc{ExOrd}: Example Ordering Perturbation). Error bars represent 95\% confidence intervals.}
\end{figure*}

\paragraph{Confounding Label Bias}
We report raw and adjusted sensitivity (label bias mitigated by CC) in Figure~\ref{fig:sensitivity_cc1}. Similar to our observations on PC, ICL becomes more sensitive when
label bias is mitigated with CC.
We also show the sensitivity scores for raw, CC and PC as table in Table~\ref{tab:sensitivity-labelbias}.

\paragraph{Sensitivity-Accuracy Correlation}
We report the correlation between prediction sensitivity and accuracy for raw and CC in Table~\ref{tab:sensitivity-corr-rawcc}. Similar to our observations on PC, there is a significant negative correlation between sensitivity and accuracy across datasets for both raw and CC.

\begin{table*}[t]
\fontsize{9.5}{9.5}\selectfont
\setlength{\tabcolsep}{2.15pt}
\centering
\begin{tabu}{c|c|c|cccccccccc|c}
\toprule
Model & Perturb Set & Label Bias & AG News & ARP & DBP & Emo & CARER & WikiQA & YAT & LYR & YRFS & MR & Avg \\
\midrule
\rowfont{\fontsize{9.5}{9.5}\selectfont}
\multirow{18}{*}{\textsc{GPT-J}} & \multirow{6}{*}{\textsc{InstH}} & \multirow{2}{*}{Raw} & $\mathbf{0.46}$ & $\mathbf{0.21}$ & $0.37$ & $0.46$ & $0.24$ & $0.01$ & $0.40$ & $0.27$ & $0.29$ & $0.20$ & $0.29$ \\
\rowfont{\fontsize{7}{7}\selectfont}
& & & $(0.12)$ & $(0.01)$ & $(0.09)$ & $(0.11)$ & $(0.17)$ & $(0.02)$ & $(0.07)$ & $(0.01)$ & $(0.08)$ & $(0.06)$ & $(0.02)$ \\
\cmidrule{3-14}
\rowfont{\fontsize{9.5}{9.5}\selectfont}
& & \multirow{2}{*}{\textsc{PC}} & $0.34$ & $0.10$ & $\mathbf{0.64}$ & $\mathbf{0.65}$ & $\mathbf{0.63}$ & $0.45$ & $\mathbf{0.64}$ & $\mathbf{0.28}$ & $\mathbf{0.44}$ & $0.14$ & $\mathbf{0.43}$ \\
\rowfont{\fontsize{7}{7}\selectfont}
&&& $(0.05)$ & $(0.05)$ & $(0.05)$ & $(0.04)$ & $(0.03)$ & $(0.01)$ & $(0.04)$ & $(0.02)$ & $(0.04)$ & $(0.04)$ & $(0.01)$ \\
\cmidrule{3-14}
\rowfont{\fontsize{9.5}{9.5}\selectfont}
&&\multirow{2}{*}{\textsc{CC}} & $0.43$ & $0.19$ & $0.42$ & $0.56$ & $0.35$ & $\mathbf{0.52}$ & $0.48$ & $0.25$ & $0.33$ & $\mathbf{0.21}$ & $0.37$ \\
\rowfont{\fontsize{7}{7}\selectfont}
&&& $(0.10)$ & $(0.08)$ & $(0.02)$ & $(0.06)$ & $(0.07)$ & $(0.05)$ & $(0.09)$ & $(0.02)$ & $(0.03)$ & $(0.08)$ & $(0.01)$ \\
\cmidrule[0.7pt]{2-14}
\rowfont{\fontsize{9.5}{9.5}\selectfont}
&\multirow{6}{*}{\textsc{InstA}} & \multirow{2}{*}{Raw} & $0.12$ & $0.05$ & $0.14$ & $0.20$ & $0.11$ & $0.01$ & $0.18$ & $\mathbf{0.10}$ & $0.18$ & $0.09$ & $0.12$ \\
\rowfont{\fontsize{7}{7}\selectfont}
&& & $(0.04)$ & $(0.01)$ & $(0.03)$ & $(0.08)$ & $(0.08)$ & $(0.01)$ & $(0.02)$ & $(0.01)$ & $(0.04)$ & $(0.03)$ & $(0.01)$ \\
\cmidrule{3-14}
\rowfont{\fontsize{9.5}{9.5}\selectfont}
&& \multirow{2}{*}{PC} & $\mathbf{0.24}$ & $0.06$ & $\mathbf{0.54}$ & $\mathbf{0.55}$ & $\mathbf{0.58}$ & $\mathbf{0.20}$ & $\mathbf{0.57}$ & $0.09$ & $\mathbf{0.38}$ & $0.08$ & $\mathbf{0.33}$ \\
\rowfont{\fontsize{7}{7}\selectfont}
&&& $(0.04)$ & $(0.02)$ & $(0.06)$ & $(0.03)$ & $(0.02)$ & $(0.01)$ & $(0.02)$ & $(0.00)$ & $(0.02)$ & $(0.02)$ & $(0.01)$ \\
\cmidrule{3-14}
\rowfont{\fontsize{9.5}{9.5}\selectfont}
&&\multirow{2}{*}{CC} & $0.13$ & $\mathbf{0.08}$ & $0.17$ & $0.27$ & $0.22$ & $0.17$ & $0.14$ & $0.09$ & $0.20$ & $\mathbf{0.11}$ & $0.16$ \\
\rowfont{\fontsize{7}{7}\selectfont}
&&& $(0.02)$ & $(0.01)$ & $(0.04)$ & $(0.04)$ & $(0.06)$ & $(0.03)$ & $(0.01)$ & $(0.02)$ & $(0.02)$ & $(0.03)$ & $(0.01)$ \\
\cmidrule[0.7pt]{2-14}
\rowfont{\fontsize{9.5}{9.5}\selectfont}
&\multirow{6}{*}{\textsc{ExOrd}} & \multirow{2}{*}{Raw} & $0.20$ & $\mathbf{0.12}$ & $0.17$ & $0.33$ & $0.12$ & $0.00$ & $0.36$ & $\mathbf{0.12}$ & $0.29$ & $0.13$ & $0.18$ \\
\rowfont{\fontsize{7}{7}\selectfont}
&& & $(0.10)$ & $(0.06)$ & $(0.07)$ & $(0.18)$ & $(0.08)$ & $(0.00)$ & $(0.18)$ & $(0.03)$ & $(0.14)$ & $(0.09)$ & $(0.01)$ \\
\cmidrule{3-14}
\rowfont{\fontsize{9.5}{9.5}\selectfont}
&& \multirow{2}{*}{PC} & $\mathbf{0.21}$ & $0.03$ & $\mathbf{0.32}$ & $\mathbf{0.52}$ & $\mathbf{0.61}$ & $\mathbf{0.16}$ & $\mathbf{0.68}$ & $0.06$ & $\mathbf{0.46}$ & $0.12$ & $\mathbf{0.32}$ \\
\rowfont{\fontsize{7}{7}\selectfont}
&&& $(0.08)$ & $(0.00)$ & $(0.05)$ & $(0.05)$ & $(0.02)$ & $(0.02)$ & $(0.03)$ & $(0.01)$ & $(0.05)$ & $(0.06)$ & $(0.01)$ \\
\cmidrule{3-14}
\rowfont{\fontsize{9.5}{9.5}\selectfont}
&&\multirow{2}{*}{CC} & $0.12$ & $0.07$ & $0.12$ & $0.46$ & $0.33$ & $0.07$ & $0.46$ & $0.10$ & $0.27$ & $\mathbf{0.24}$ & $0.23$ \\
\rowfont{\fontsize{7}{7}\selectfont}
&&& $(0.03)$ & $(0.03)$ & $(0.05)$ & $(0.08)$ & $(0.07)$ & $(0.06)$ & $(0.08)$ & $(0.01)$ & $(0.07)$ & $(0.08)$ & $(0.03)$\\
\midrule
\rowfont{\fontsize{9.5}{9.5}\selectfont}
\multirow{18}{*}{\textsc{GPT-Neo}} & \multirow{6}{*}{\textsc{InstH}} & \multirow{2}{*}{Raw} & $\mathbf{0.48}$ & $0.25$ & $0.42$ & $0.51$ & $0.14$ & $0.00$ & $0.40$ & $\mathbf{0.36}$ & $0.38$ & $0.18$ & $0.31$\\
\rowfont{\fontsize{7}{7}\selectfont}
&&& $(0.10)$ & $(0.14)$ & $(0.10)$ & $(0.18)$ & $(0.19)$ & $(0.00)$ & $(0.26)$ & $(0.06)$ & $(0.11)$ & $(0.15)$ & $(0.05)$\\
\cmidrule{3-14}
\rowfont{\fontsize{9.5}{9.5}\selectfont}
& & \multirow{2}{*}{\textsc{PC}} & $0.43$ & $0.31$ & $\mathbf{0.72}$ & $\mathbf{0.68}$ & $\mathbf{0.64}$ & $0.36$ & $\mathbf{0.81}$ & $0.31$ & $\mathbf{0.55}$ & $0.21$ & $\mathbf{0.50}$\\
\rowfont{\fontsize{7}{7}\selectfont}
&&& $(0.04)$ & $(0.06)$ & $(0.09)$ & $(0.02)$ & $(0.04)$ & $(0.01)$ & $(0.02)$ & $(0.02)$ & $(0.03)$ & $(0.06)$ & $(0.01)$\\
\cmidrule{3-14}
\rowfont{\fontsize{9.5}{9.5}\selectfont}
&&\multirow{2}{*}{\textsc{CC}} & $0.45$ & $\mathbf{0.36}$ & $0.60$ & $0.54$ & $0.37$ & $\mathbf{0.46}$ & $0.62$ & $0.35$ & $0.41$ & $\mathbf{0.29}$ & $0.45$\\
\rowfont{\fontsize{7}{7}\selectfont}
&&& $(0.10)$ & $(0.05)$ & $(0.03)$ & $(0.08)$ & $(0.05)$ & $(0.04)$ & $(0.07)$ & $(0.02)$ & $(0.05)$ & $(0.12)$ & $(0.02)$\\
\cmidrule[0.7pt]{2-14}
\rowfont{\fontsize{9.5}{9.5}\selectfont}
&\multirow{6}{*}{\textsc{InstA}} & \multirow{2}{*}{Raw} & $0.11$ & $0.07$ & $0.17$ & $0.25$ & $0.08$ & $0.00$ & $0.15$ & $0.12$ & $0.21$ & $0.08$ & $0.12$\\
\rowfont{\fontsize{7}{7}\selectfont}
&&& $(0.03)$ & $(0.03)$ & $(0.08)$ & $(0.11)$ & $(0.10)$ & $(0.00)$ & $(0.07)$ & $(0.01)$ & $(0.08)$ & $(0.06)$ & $(0.01)$\\
\cmidrule{3-14}
\rowfont{\fontsize{9.5}{9.5}\selectfont}
&& \multirow{2}{*}{PC} & $\mathbf{0.27}$ & $0.10$ & $\mathbf{0.61}$ & $\mathbf{0.58}$ & $\mathbf{0.57}$ & $0.17$ & $\mathbf{0.64}$ & $0.14$ & $\mathbf{0.51}$ & $0.12$ & $\mathbf{0.37}$\\
\rowfont{\fontsize{7}{7}\selectfont}
&&& $(0.03)$ & $(0.02)$ & $(0.09)$ & $(0.02)$ & $(0.04)$ & $(0.01)$ & $(0.02)$ & $(0.02)$ & $(0.03)$ & $(0.03)$ & $(0.01)$\\
\cmidrule{3-14}
\rowfont{\fontsize{9.5}{9.5}\selectfont}
&&\multirow{2}{*}{CC} & $0.11$ & $\mathbf{0.11}$ & $0.20$ & $0.28$ & $0.24$ & $\mathbf{0.21}$ & $0.20$ & $\mathbf{0.14}$ & $0.29$ & $\mathbf{0.13}$ & $0.19$\\
\rowfont{\fontsize{7}{7}\selectfont}
&&& $(0.01)$ & $(0.02)$ & $(0.02)$ & $(0.05)$ & $(0.03)$ & $(0.03)$ & $(0.03)$ & $(0.02)$ & $(0.02)$ & $(0.04)$ & $(0.01)$\\
\cmidrule[0.7pt]{2-14}
\rowfont{\fontsize{9.5}{9.5}\selectfont}
&\multirow{6}{*}{\textsc{ExOrd}} & \multirow{2}{*}{Raw} & $0.16$ & $\mathbf{0.16}$ & $0.16$ & $0.21$ & $0.12$ & $0.00$ & $0.31$ & $\mathbf{0.28}$ & $0.37$ & $\mathbf{0.36}$ & $0.21$\\
\rowfont{\fontsize{7}{7}\selectfont}
&&& $(0.07)$ & $(0.09)$ & $(0.02)$ & $(0.19)$ & $(0.14)$ & $(0.00)$ & $(0.19)$ & $(0.04)$ & $(0.15)$ & $(0.18)$ & $(0.04)$\\
\cmidrule{3-14}
\rowfont{\fontsize{9.5}{9.5}\selectfont}
&& \multirow{2}{*}{PC} & 
$\mathbf{0.23}$ & $0.08$ & $\mathbf{0.44}$ & $\mathbf{0.57}$ & $\mathbf{0.65}$ & $0.10$ & $\mathbf{0.70}$ & $0.17$ & $\mathbf{0.62}$ & $0.23$ & $\mathbf{0.38}$\\
\rowfont{\fontsize{7}{7}\selectfont}
&&& $(0.07)$ & $(0.02)$ & $(0.05)$ & $(0.05)$ & $(0.03)$ & $(0.01)$ & $(0.04)$ & $(0.04)$ & $(0.10)$ & $(0.07)$ & $(0.02)$\\
\cmidrule{3-14}
\rowfont{\fontsize{9.5}{9.5}\selectfont}
&&\multirow{2}{*}{CC} & $0.11$ & $0.14$ & $0.23$ & $0.39$ & $0.33$ & $\mathbf{0.17}$ & $0.41$ & $0.20$ & $0.43$ & $0.27$ & $0.27$\\
\rowfont{\fontsize{7}{7}\selectfont}
&&& $(0.02)$ & $(0.06)$ & $(0.03)$ & $(0.02)$ & $(0.07)$ & $(0.04)$ & $(0.07)$ & $(0.05)$ & $(0.12)$ & $(0.14)$ & $(0.02)$\\
\bottomrule
\end{tabu}
\caption{\label{tab:sensitivity-labelbias}
We compare the raw sensitivity with the adjusted sensitivity after mitigating label bias. We observe that the adjusted sensitivity is consistently higher than the raw sensitivity for all three perturbation sets (\textsc{InstH}: Human Instruction Perturbation, \textsc{InstA}: Automatic Instruction Perturbation, and \textsc{ExOrd}: Example Ordering Perturbation). The standard deviation across five randomly sampled sets of few-shot examples is reported in parenthesis. 
}
\end{table*}

\subsection{Sensitivity-Based Selective Few-shot Prediction}
\label{sec:additional-sensel-results}

\begin{figure*}[t]
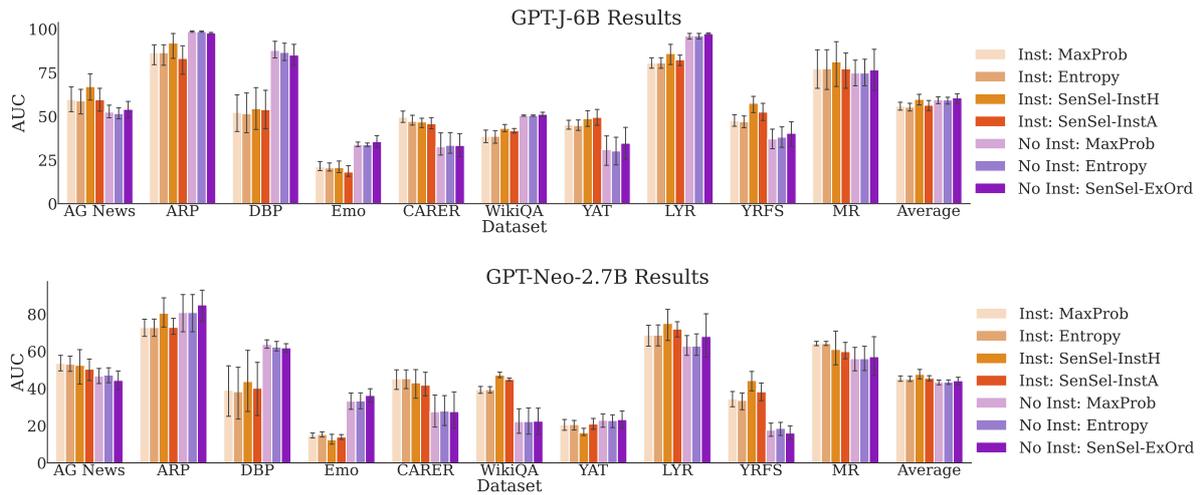

\centering
\includegraphics[scale=0.0325]{figures/sensel_auc_gpt-j-6B-cc.pdf}\\
\vspace{1em}
\includegraphics[scale=0.0325]{figures/sensel_auc_gpt-neo-2.7B-cc.pdf}\\
\caption{\label{fig:auc1} We compare our \textsc{SenSel} method (confounding label bias mitigated by CC) to the \textsc{MaxProb} baseline. \textsc{SenSel} consistently outperforms \textsc{MaxProb} under both the \textsc{Inst} setting and the \textsc{No Inst} setting.}
\end{figure*}

Similar to results on PC, all three variants of
\textsc{SenSel} consistently outperform both \textsc{MaxProb} and \textsc{Entropy} when CC is used to mitigate label bias (Figure~\ref{fig:auc1}). Among the three perturbation sets, \textsc{SenSel} with human-written instruction perturbations performs the best (outperforming \textsc{MaxProb} and \textsc{Entropy} by $\mathbf{+3.9}$ AUC points on \textsc{GPT-J-6B} and $\mathbf{+0.8}$ AUC points on \textsc{GPT-Neo-2.7B}). Similar to results on PC, \textsc{SenSel-ExOrd} outperforms \textsc{MaxProb} and \textsc{Entropy} consistently even when instructions are not available. We also show the AUC scores as table in Table~\ref{tab:auc-ccpc-gpt-j},\ref{tab:auc-ccpc-gptneo}.

\begin{figure*}[t]
\centering
\includegraphics[scale=0.028]{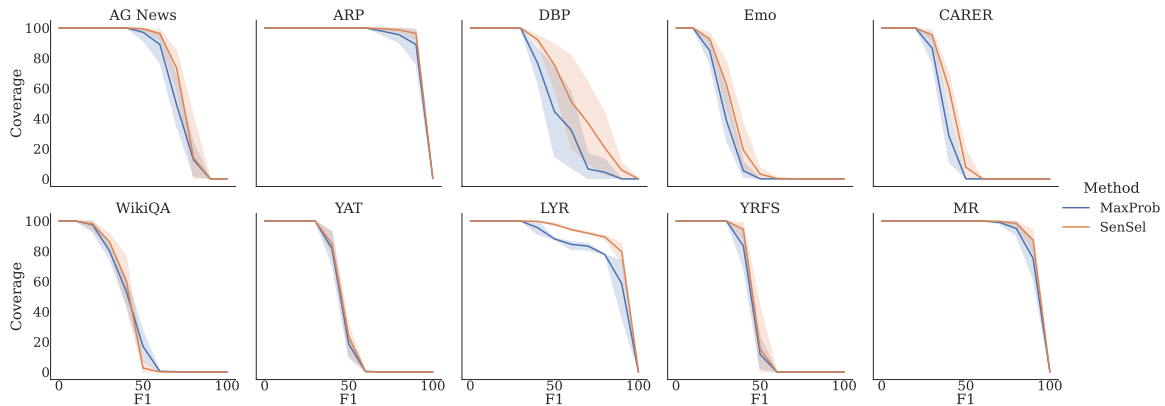}
\caption{\label{fig:coverage_curves} We plot the Coverage-F1 curves of \textsc{MaxProb} and \textsc{SenSel-InstH} of \textsc{GPT-J-6B} (confounding label bias mitigated by PC). \textsc{SenSel-InstH} consistently achieves higher coverage rates at different F1 thresholds compared to \textsc{MaxProb}. Color bands represent 95\% confidence intervals.}
\end{figure*}

We also plot the Coverage-F1 curves, which show coverage rates at different F1 thresholds (Figure~\ref{fig:coverage_curves}). The coverage-F1 curves for \textsc{SenSel-InstH} and \textsc{MaxProb} further verify that \textsc{SenSel} consistently outperforms \textsc{MaxProb} on different thresholds (Figure~\ref{fig:coverage_curves}).

\begin{table*}[t]
\fontsize{8}{8}\selectfont
\setlength{\tabcolsep}{1.8pt}
\centering
\begin{tabu}{c|c|c|cccccccccc|c}
\toprule
Model & Label Bias & Perturb & AG News & ARP & DBP & Emo & CARER & WikiQA & YAT & LYR & YRFS & MR & Avg \\
\midrule
\rowfont{\fontsize{9}{9}\selectfont}
\multirow{12}{*}{\textsc{GPT-J-6B}} & \multirow{6}{*}{Raw} & \multirow{2}{*}{\textsc{InstH}} & $-0.49$ & $-0.50$ & $-0.11$ & $-0.21$ & $-0.12$ & $-0.09$ & $-0.25$ & $-0.54$ & $-0.24$ & $-0.31$ & $-0.29$ \\
\rowfont{\fontsize{7}{7}\selectfont}
&& & $(0.14)$ & $(0.10)$ & $(0.17)$ & $(0.13)$ & $(0.12)$ & $(0.05)$ & $(0.03)$ & $(0.04)$ & $(0.05)$ & $(0.13)$ & $(0.04)$ \\
\cmidrule{3-14}
\rowfont{\fontsize{9}{9}\selectfont}
&& \multirow{2}{*}{\textsc{InstA}} & $-0.24$ & $-0.29$ & $-0.17$ & $-0.09$ & $-0.06$ & $-0.32$ & $-0.19$ & $-0.31$ & $-0.12$ & $-0.18$ & $-0.20$ \\
\rowfont{\fontsize{7}{7}\selectfont}
&&& $(0.08)$ & $(0.12)$ & $(0.16)$ & $(0.12)$ & $(0.10)$ & $(0.23)$ & $(0.13)$ & $(0.03)$ & $(0.07)$ & $(0.11)$ & $(0.06)$ \\
\cmidrule{3-14}
\rowfont{\fontsize{9}{9}\selectfont}
&&\multirow{2}{*}{\textsc{ExOrd}} & $-0.14$ & $-0.36$ & $-0.16$ & $-0.30$ & $-0.08$ & $/$ & $-0.13$ & $-0.59$ & $-0.22$ & $-0.32$ & $-0.26$ \\
\rowfont{\fontsize{7}{7}\selectfont}
&&& $(0.12)$ & $(0.16)$ & $(0.22)$ & $(0.19)$ & $(0.04)$ & $/$ & $(0.10)$ & $(0.03)$ & $(0.07)$ & $(0.13)$ & $(0.05)$ \\
\cmidrule[0.7pt]{2-14}
\rowfont{\fontsize{9}{9}\selectfont}
&\multirow{6}{*}{CC} & \multirow{2}{*}{\textsc{InstH}} & $-0.50$ & $-0.57$ & $-0.38$ & $-0.06$ & $-0.29$ & $-0.34$ & $-0.35$ & $-0.48$ & $-0.28$ & $-0.48$ & $-0.37$ \\
\rowfont{\fontsize{7}{7}\selectfont}
&& & $(0.07)$ & $(0.05)$ & $(0.09)$ & $(0.04)$ & $(0.02)$ & $(0.09)$ & $(0.02)$ & $(0.10)$ & $(0.02)$ & $(0.10)$ & $(0.02)$ \\
\cmidrule{3-14}
\rowfont{\fontsize{9}{9}\selectfont}
&& \multirow{2}{*}{\textsc{InstA}} & $-0.26$ & $-0.24$ & $-0.38$ & $0.00$ & $-0.14$ & $-0.35$ & $-0.28$ & $-0.33$ & $-0.20$ & $-0.38$ & $-0.26$ \\
\rowfont{\fontsize{7}{7}\selectfont}
&&& $(0.05)$ & $(0.12)$ & $(0.08)$ & $(0.03)$ & $(0.05)$ & $(0.03)$ & $(0.04)$ & $(0.09)$ & $(0.03)$ & $(0.08)$ & $(0.02)$ \\
\cmidrule{3-14}
\rowfont{\fontsize{9}{9}\selectfont}
&&\multirow{2}{*}{\textsc{ExOrd}} & $-0.19$ & $-0.47$ & $-0.52$ & $-0.22$ & $-0.30$ & $-0.33$ & $-0.37$ & $-0.58$ & $-0.20$ & $-0.47$ & $-0.37$ \\
\rowfont{\fontsize{7}{7}\selectfont}
&&& $(0.11)$ & $(0.07)$ & $(0.03)$ & $(0.08)$ & $(0.05)$ & $(0.09)$ & $(0.06)$ & $(0.05)$ & $(0.03)$ & $(0.05)$ & $(0.02)$ \\
\midrule
\rowfont{\fontsize{9}{9}\selectfont}
\multirow{12}{*}{\textsc{GPT-Neo-2.7B}} & \multirow{6}{*}{Raw} & \multirow{2}{*}{\textsc{InstH}} & $-0.35$ & $-0.37$ & $-0.07$ & $-0.16$ & $/$ & $-0.07$ & $-0.06$ & $-0.45$ & $-0.15$ & $-0.28$ & $-0.22$\\
\rowfont{\fontsize{7}{7}\selectfont}
&& & $(0.17)$ & $(0.24)$ & $(0.17)$ & $(0.19)$ & $/$ & $(0.00)$ & $(0.16)$ & $(0.16)$ & $(0.10)$ & $(0.11)$ & $(0.04)$\\
\cmidrule{3-14}
\rowfont{\fontsize{9}{9}\selectfont}
&& \multirow{2}{*}{\textsc{InstA}} & $-0.14$ & $-0.19$ & $-0.18$ & $-0.10$ & $-0.05$ & $-0.16$ & $-0.14$ & $-0.11$ & $-0.02$ & $-0.13$ & $-0.12$\\
\rowfont{\fontsize{7}{7}\selectfont}
&&& $(0.03)$ & $(0.10)$ & $(0.09)$ & $(0.10)$ & $(0.08)$ & $(0.03)$ & $(0.10)$ & $(0.05)$ & $(0.08)$ & $(0.12)$ & $(0.03)$\\
\cmidrule{3-14}
\rowfont{\fontsize{9}{9}\selectfont}
&&\multirow{2}{*}{\textsc{ExOrd}} &$-0.12$ & $/$ & $-0.11$ & $-0.25$ & $-0.09$ & $-0.07$ & $-0.11$ & $-0.45$ & $-0.10$ & $-0.29$ & $-0.18$\\
\rowfont{\fontsize{7}{7}\selectfont}
&&& $(0.14)$ & $/$ & $(0.23)$ & $(0.14)$ & $(0.12)$ & $(0.03)$ & $(0.11)$ & $(0.07)$ & $(0.09)$ & $(0.21)$ & $(0.02)$\\
\cmidrule[0.7pt]{2-14}
\rowfont{\fontsize{9}{9}\selectfont}
&\multirow{6}{*}{CC} & \multirow{2}{*}{\textsc{InstH}} & $-0.37$ & $-0.61$ & $-0.48$ & $0.04$ & $-0.33$ & $-0.33$ & $-0.18$ & $-0.47$ & $-0.25$ & $-0.36$ & $-0.33$\\
\rowfont{\fontsize{7}{7}\selectfont}
&& & $(0.13)$ & $(0.04)$ & $(0.15)$ & $(0.09)$ & $(0.05)$ & $(0.05)$ & $(0.05)$ & $(0.11)$ & $(0.06)$ & $(0.15)$ & $(0.02)$\\
\cmidrule{3-14}
\rowfont{\fontsize{9}{9}\selectfont}
&& \multirow{2}{*}{\textsc{InstA}} & $-0.20$ & $-0.26$ & $-0.31$ & $-0.04$ & $-0.21$ & $-0.38$ & $-0.23$ & $-0.20$ & $-0.13$ & $-0.26$ & $-0.22$\\
\rowfont{\fontsize{7}{7}\selectfont}
&&& $(0.07)$ & $(0.05)$ & $(0.12)$ & $(0.09)$ & $(0.05)$ & $(0.04)$ & $(0.06)$ & $(0.05)$ & $(0.06)$ & $(0.07)$ & $(0.02)$\\
\cmidrule{3-14}
\rowfont{\fontsize{9}{9}\selectfont}
&&\multirow{2}{*}{\textsc{ExOrd}} & $-0.10$ & $-0.44$ & $-0.43$ & $-0.34$ & $-0.25$ & $0.21$ & $-0.21$ & $-0.35$ & $-0.01$ & $-0.31$ & $-0.22$\\
\rowfont{\fontsize{7}{7}\selectfont}
&&& $(0.09)$ & $(0.05)$ & $(0.05)$ & $(0.03)$ & $(0.09)$ & $(0.11)$ & $(0.09)$ & $(0.07)$ & $(0.04)$ & $(0.16)$ & $(0.03)$\\
\bottomrule
\end{tabu}
\caption{\label{tab:sensitivity-corr-rawcc} We report the Pearson correlation coefficient (and its standard deviation in parenthesis) between ICL sensitivity and accuracy across five randomly sampled sets
of few-shot examples (label bias mitigated with CC). 
We observe a strong negative correlation between the ICL sensitivity and accuracy for all three perturbation sets.  
}
\end{table*}

\begin{table*}[t]
\fontsize{9.5}{9.5}\selectfont
\setlength{\tabcolsep}{1.7pt}
\centering
\begin{tabu}{c|c|c|cccccccccc|c}
\toprule
\rowfont{\fontsize{9.5}{9.5}\selectfont}
Label Bias & Setting & Method & AG News & ARP &  DBP & Emo & CARER & WikiQA & YAT & LYR & YRFS & MR & Avg \\
\midrule
\rowfont{\fontsize{9.5}{9.5}\selectfont}
\multirow{14}{*}{\textsc{PC}} & \multirow{8}{*}{\textsc{Inst}} & \multirow{2}{*}{\textsc{MaxProb}} & $64.9$ & $94.2$ & $51.0$ & $27.2$ & $36.1$ & $39.7$ & $43.9$ & $82.7$ & $42.6$ & $92.3$ & $57.5$ \\
\rowfont{\fontsize{7}{7}\selectfont}
&&& $(4.0)$ & $(3.2)$ & $(8.5)$ & $(2.2)$ & $(2.9)$ & $(2.2)$ & $(1.3)$ & $(1.9)$ & $(1.6)$ & $(2.2)$ & $(0.7)$ \\
\cmidrule{3-14}  
\rowfont{\fontsize{9.5}{9.5}\selectfont}
&&\multirow{2}{*}{\textsc{Entropy}} & $63.7$ & $89.8$ & $51.1$ & $27.3$ & $36.1$ & $40.2$ & $43.7$ & $77.6$ & $42.4$ & $91.9$ & $56.4$\\
\rowfont{\fontsize{7}{7}\selectfont}
&&& $(3.6)$ & $(2.9)$ & $(8.9)$ & $(2.3)$ & $(2.9)$ & $(1.9)$ & $(1.4)$ & $(5.0)$ & $(1.6)$ & $(2.2)$ & $(0.9)$\\
\cmidrule{3-14}
\rowfont{\fontsize{9.5}{9.5}\selectfont}
&&\multirow{2}{*}{\textsc{SenSel-InstH}} & $\mathbf{70.2}$ & $\mathbf{96.6}$ & $63.1$ & $\mathbf{32.7}$ & $\mathbf{40.9}$ & $37.7$ & $45.1$ & $\mathbf{91.0}$ & $\mathbf{45.2}$ & $\mathbf{93.3}$ & $\mathbf{61.6}$ \\
\rowfont{\fontsize{7}{7}\selectfont}
&&&$(4.5)$ & $(1.8)$ & $(11.0)$ & $(4.3)$ & $(2.1)$ & $(4.8)$ & $(1.5)$ & $(3.6)$ & $(3.5)$ & $(1.1)$ & $(1.1)$ \\
\cmidrule{3-14}
\rowfont{\fontsize{9.5}{9.5}\selectfont}
&&\multirow{2}{*}{\textsc{SenSel-InstA}} & $64.6$ & $93.3$ & $\mathbf{65.6}$ & $30.8$ & $39.2$ & $\mathbf{42.1}$ & $\mathbf{47.2}$ & $85.1$ & $42.9$ & $90.8$ & $60.2$ \\
\rowfont{\fontsize{7}{7}\selectfont}
&&& $(3.5)$ & $(3.0)$ & $(10.5)$ & $(3.5)$ & $(2.5)$ & $(2.4)$ & $(1.5)$ & $(1.3)$ & $(1.8)$ & $(1.8)$ & $(0.8)$ \\
\cmidrule[0.7pt]{2-14}
\rowfont{\fontsize{9.5}{9.5}\selectfont}
&\multirow{6}{*}{\textsc{No Inst}} & \multirow{2}{*}{\textsc{MaxProb}} & $65.7$ & $97.5$ & $76.7$ & $25.2$ & $\mathbf{27.7}$ & $43.3$ & $27.7$ & $94.5$ & $38.3$ & $91.9$ & $58.8$ \\
\rowfont{\fontsize{7}{7}\selectfont}
&&& $(6.0)$ & $(0.4)$ & $(6.1)$ & $(2.4)$ & $(4.9)$ & $(2.5)$ & $(6.6)$ & $(1.4)$ & $(2.9)$ & $(3.4)$ & $(1.5)$ \\
\cmidrule{3-14} 
\rowfont{\fontsize{9.5}{9.5}\selectfont}
&&\multirow{2}{*}{\textsc{Entropy}} & $62.7$ & $94.3$ & $77.2$ & $25.4$ & $27.6$ & $42.8$ & $27.7$ & $88.8$ & $37.7$ & $90.6$ & $57.5$\\
\rowfont{\fontsize{7}{7}\selectfont}
&&& $(5.5)$ & $(6.1)$ & $(7.5)$ & $(2.5)$ & $(4.9)$ & $(1.9)$ & $(6.6)$ & $(4.2)$ & $(3.1)$ & $(3.4)$ & $(1.9)$\\
\cmidrule{3-14}
\rowfont{\fontsize{9.5}{9.5}\selectfont}
&&\multirow{2}{*}{\textsc{SenSel-ExOrd}} & $\mathbf{65.8}$ & $\mathbf{97.6}$ & $\mathbf{92.2}$ & $\mathbf{25.9}$ & $31.1$ & $\mathbf{44.1}$ & $\mathbf{31.5}$ & $\mathbf{97.3}$ & $\mathbf{40.4}$ & $\mathbf{92.3}$ & $\mathbf{61.8}$ \\
\rowfont{\fontsize{7}{7}\selectfont}
&&& $(5.1)$ & $(0.2)$ & $(3.9)$ & $(2.1)$ & $(6.4)$ & $(3.2)$ & $(7.9)$ & $(0.5)$ & $(3.3)$ & $(2.6)$ & $(2.1)$ \\
\midrule[0.7pt]
\rowfont{\fontsize{9.5}{9.5}\selectfont}
\multirow{14}{*}{\textsc{CC}} & \multirow{8}{*}{\textsc{Inst}} & \multirow{2}{*}{\textsc{MaxProb}} & $59.6$ & $86.5$ & $52.3$ & $\mathbf{21.2}$ & $\mathbf{49.5}$ & $38.6$ & $45.2$ & $80.6$ & $47.6$ & $77.3$ & $55.8$ \\
\rowfont{\fontsize{7}{7}\selectfont}
&&& $(7.7)$ & $(7.1)$ & $(12.7)$ & $(3.1)$ & $(3.8)$ & $(4.2)$ & $(2.9)$ & $(3.4)$ & $(4.0)$ & $(13.5)$ & $(2.6)$ \\
\cmidrule{3-14} 
\rowfont{\fontsize{9.5}{9.5}\selectfont}
&&\multirow{2}{*}{\textsc{Entropy}} & $58.9$ & $86.5$ & $51.6$ & $20.5$ & $47.2$ & $38.6$ & $44.8$ & $80.6$ & $46.9$ & $77.3$ & $55.3$\\
\rowfont{\fontsize{7}{7}\selectfont}
&&& $(8.2)$ & $(7.1)$ & $(12.8)$ & $(2.8)$ & $(3.2)$ & $(4.2)$ & $(3.4)$ & $(3.5)$ & $(3.8)$ & $(13.5)$ & $(2.5)$\\
\cmidrule{3-14}
\rowfont{\fontsize{9.5}{9.5}\selectfont}
&&\multirow{2}{*}{\textsc{SenSel-InstH}} & $\mathbf{67.1}$ & $\mathbf{92.1}$ & $\mathbf{54.4}$ & $20.6$ & $46.7$ & $\mathbf{43.3}$ & $\mathbf{49.3}$ & $\mathbf{85.8}$ & $\mathbf{57.2}$ & $\mathbf{81.0}$ & $\mathbf{59.7}$ \\
\rowfont{\fontsize{7}{7}\selectfont}
&&& $(8.6)$ & $(8.6)$ & $(14.0)$ & $(3.7)$ & $(3.4)$ & $(2.4)$ & $(5.8)$ & $(6.5)$ & $(5.8)$ & $(15.9)$ & $(3.7)$ \\
\cmidrule{3-14} 
\rowfont{\fontsize{9.5}{9.5}\selectfont}
&&\multirow{2}{*}{\textsc{SenSel-InstA}} & $59.5$ & $83.5$ & $53.8$ & $18.1$ & $45.4$ & $41.9$ & $\mathbf{49.3}$ & $82.4$ & $52.5$ & $77.4$ & $56.4$ \\
\rowfont{\fontsize{7}{7}\selectfont}
&&& $(7.9)$ & $(9.8)$ & $(12.7)$ & $(3.2)$ & $(4.0)$ & $(1.9)$ & $(5.7)$ & $(3.4)$ & $(5.6)$ & $(12.3)$ & $(3.2)$ \\
\cmidrule[0.7pt]{2-14}
\rowfont{\fontsize{9.5}{9.5}\selectfont}
&\multirow{6}{*}{\textsc{No Inst}} &\multirow{2}{*}{\textsc{MaxProb}} & $51.4$ & $94.7$ & $\mathbf{87.0}$ & $31.3$ & $32.1$ & $50.7$ & $27.7$ & $93.0$ & $37.7$ & $73.3$ & $57.9$ \\
\rowfont{\fontsize{7}{7}\selectfont}
&&& $(7.7)$ & $(2.4)$ & $(5.8)$ & $(4.4)$ & $(6.4)$ & $(0.7)$ & $(7.9)$ & $(3.9)$ & $(6.9)$ & $(7.7)$ & $(2.8)$ \\
\cmidrule{3-14} 
\rowfont{\fontsize{9.5}{9.5}\selectfont}
&&\multirow{2}{*}{\textsc{Entropy}} & $51.7$ & $\mathbf{98.5}$ & $86.6$ & $33.8$ & $\mathbf{33.4}$ & $50.5$ & $\mathbf{30.2}$ & $96.2$ & $38.1$ & $74.9$ & $59.4$\\
\rowfont{\fontsize{7}{7}\selectfont}
&&& $(3.7)$ & $(0.4)$ & $(5.7)$ & $(1.1)$ & $(7.4)$ & $(0.4)$ & $(9.0)$ & $(1.8)$ & $(6.9)$ & $(8.7)$ & $(2.3)$\\
\cmidrule{3-14}
\rowfont{\fontsize{9.5}{9.5}\selectfont}
&&\multirow{2}{*}{\textsc{SenSel-ExOrd}} & $\mathbf{52.9}$ & $96.4$ & $83.2$ & $\mathbf{34.5}$ & $33.1$ & $\mathbf{51.1}$ & $29.5$ & $\mathbf{97.1}$ & $\mathbf{40.9}$ & $\mathbf{80.8}$ & $\mathbf{60.0}$ \\
\rowfont{\fontsize{7}{7}\selectfont}
&&& $(11.0)$ & $(1.8)$ & $(7.3)$ & $(5.7)$ & $(7.4)$ & $(1.3)$ & $(7.6)$ & $(1.6)$ & $(9.3)$ & $(6.2)$ & $(2.2)$ \\
\bottomrule
\end{tabu}
\caption{\label{tab:auc-ccpc-gpt-j} We compare our \textsc{SenSel} method to the \textsc{MaxProb} baseline and the \textsc{Entropy} baseline on the \textsc{GPT-J-6B} model. \textsc{SenSel} consistently outperforms both baselines under both the \textsc{Inst} setting and the \textsc{No Inst} setting. The standard deviation across five randomly sampled sets of few-shot examples is reported in parenthesis. 
}
\end{table*}

\begin{table*}[t]
\fontsize{9.5}{9.5}\selectfont
\setlength{\tabcolsep}{1.7pt}
\centering
\begin{tabu}{c|c|c|cccccccccc|c}
\toprule
\rowfont{\fontsize{9.5}{9.5}\selectfont}
Label Bias & Setting & Method & AG News & ARP &  DBP & Emo & CARER & WikiQA & YAT & LYR & YRFS & MR & Avg \\
\midrule
\rowfont{\fontsize{9.5}{9.5}\selectfont}
\multirow{14}{*}{\textsc{PC}} & \multirow{8}{*}{\textsc{Inst}} & \multirow{2}{*}{\textsc{MaxProb}} & $60.7$ & $80.0$ & $43.0$ & $20.3$ & $35.8$ & $45.5$ & $26.9$ & $73.0$ & $29.7$ & $86.1$ & $50.1$\\
\rowfont{\fontsize{7}{7}\selectfont}
&&& $(3.6)$ & $(6.4)$ & $(13.3)$ & $(1.8)$ & $(3.9)$ & $(1.0)$ & $(1.9)$ & $(3.7)$ & $(3.1)$ & $(3.7)$ & $(1.4)$\\
\cmidrule{3-14}  
\rowfont{\fontsize{9.5}{9.5}\selectfont}
&&\multirow{2}{*}{\textsc{Entropy}} & $60.2$ & $77.8$ & $42.6$ & $20.3$ & $35.8$ & $45.2$ & $26.6$ & $65.2$ & $29.4$ & $85.6$ & $48.9$\\
\rowfont{\fontsize{7}{7}\selectfont}
&&& $(3.5)$ & $(6.4)$ & $(13.2)$ & $(1.8)$ & $(3.9)$ & $(0.9)$ & $(2.0)$ & $(4.6)$ & $(3.0)$ & $(3.3)$ & $(1.3)$\\
\cmidrule{3-14}
\rowfont{\fontsize{9.5}{9.5}\selectfont}
&&\multirow{2}{*}{\textsc{SenSel-InstH}} & $\mathbf{67.0}$ & $\mathbf{86.4}$ & $52.6$ & $\mathbf{22.0}$ & $\mathbf{40.4}$ & $\mathbf{51.9}$ & $27.1$ & $\mathbf{84.1}$ & $\mathbf{32.0}$ & $\mathbf{87.5}$ & $\mathbf{55.1}$\\
\rowfont{\fontsize{7}{7}\selectfont}
&&& $(4.9)$ & $(7.4)$ & $(16.2)$ & $(3.8)$ & $(3.9)$ & $(1.5)$ & $(2.3)$ & $(5.8)$ & $(4.2)$ & $(3.5)$ & $(1.7)$\\
\cmidrule{3-14}
\rowfont{\fontsize{9.5}{9.5}\selectfont}
&&\multirow{2}{*}{\textsc{SenSel-InstA}} & $60.3$ & $74.5$ & $\mathbf{54.9}$ & $21.2$ & $38.8$ & $45.4$ & $\mathbf{29.4}$ & $75.0$ & $30.2$ & $82.9$ & $51.3$\\
\rowfont{\fontsize{7}{7}\selectfont}
&&& $(2.9)$ & $(6.5)$ & $(16.0)$ & $(2.9)$ & $(4.3)$ & $(1.1)$ & $(1.4)$ & $(3.6)$ & $(4.6)$ & $(3.9)$ & $(1.8)$\\
\cmidrule[0.7pt]{2-14}
\rowfont{\fontsize{9.5}{9.5}\selectfont}
&\multirow{6}{*}{\textsc{No Inst}} & \multirow{2}{*}{\textsc{MaxProb}} & $\mathbf{62.8}$ & $94.8$ & $62.5$ & $22.3$ & $24.7$ & $51.4$ & $16.8$ & $84.4$ & $23.9$ & $\mathbf{69.7}$ & $51.3$\\
\rowfont{\fontsize{7}{7}\selectfont}
&&& $(3.3)$ & $(0.7)$ & $(7.0)$ & $(2.8)$ & $(5.3)$ & $(1.0)$ & $(4.4)$ & $(6.8)$ & $(2.9)$ & $(7.5)$ & $(2.0)$\\
\cmidrule{3-14} 
\rowfont{\fontsize{9.5}{9.5}\selectfont}
&&\multirow{2}{*}{\textsc{Entropy}} & $59.7$ & $92.7$ & $60.7$ & $22.5$ & $24.7$ & $50.2$ & $16.8$ & $80.2$ & $23.8$ & $69.5$ & $50.1$\\
\rowfont{\fontsize{7}{7}\selectfont}
&&& $(3.5)$ & $(3.5)$ & $(7.9)$ & $(2.8)$ & $(5.3)$ & $(1.0)$ & $(4.5)$ & $(6.8)$ & $(2.8)$ & $(7.8)$ & $(1.9)$\\
\cmidrule{3-14}
\rowfont{\fontsize{9.5}{9.5}\selectfont}
&&\multirow{2}{*}{\textsc{SenSel-ExOrd}} & $56.6$ & $\mathbf{95.3}$ & $\mathbf{76.8}$ & $\mathbf{24.4}$ & $\mathbf{28.0}$ & $\mathbf{53.0}$ & $\mathbf{17.4}$ & $\mathbf{87.0}$ & $\mathbf{25.5}$ & $62.7$ & $\mathbf{52.7}$\\
\rowfont{\fontsize{7}{7}\selectfont}
&&& $(3.0)$ & $(1.4)$ & $(6.2)$ & $(3.5)$ & $(6.7)$ & $(0.7)$ & $(6.1)$ & $(8.2)$ & $(3.8)$ & $(8.2)$ & $(2.3)$\\
\midrule[0.7pt]
\rowfont{\fontsize{9.5}{9.5}\selectfont}
\multirow{14}{*}{\textsc{CC}} & \multirow{8}{*}{\textsc{Inst}} & \multirow{2}{*}{\textsc{MaxProb}} & $\mathbf{53.5}$ & $72.8$ & $39.1$ & $14.8$ & $45.3$ & $39.4$ & $20.5$ & $68.7$ & $34.5$ & $\mathbf{64.2}$ & $45.3$\\
\rowfont{\fontsize{7}{7}\selectfont}
&&& $(4.9)$ & $(5.4)$ & $(16.7)$ & $(1.6)$ & $(5.8)$ & $(2.1)$ & $(3.2)$ & $(6.4)$ & $(4.5)$ & $(1.3)$ & $(1.7)$\\
\cmidrule{3-14} 
\rowfont{\fontsize{9.5}{9.5}\selectfont}
&&\multirow{2}{*}{\textsc{Entropy}} & $53.1$ & $72.8$ & $38.3$ & $\mathbf{15.3}$ & $\mathbf{45.3}$ & $39.4$ & $20.6$ & $68.7$ & $33.6$ & $64.2$ & $45.1$\\
\rowfont{\fontsize{7}{7}\selectfont}
&&& $(4.8)$ & $(5.4)$ & $(16.0)$ & $(1.7)$ & $(5.5)$ & $(2.1)$ & $(3.1)$ & $(6.4)$ & $(5.0)$ & $(1.3)$ & $(1.6)$\\
\cmidrule{3-14}
\rowfont{\fontsize{9.5}{9.5}\selectfont}
&&\multirow{2}{*}{\textsc{SenSel-InstH}} &$52.5$ & $\mathbf{80.5}$ & $\mathbf{43.8}$ & $12.5$ & $43.1$ & $\mathbf{47.2}$ & $16.3$ & $\mathbf{75.0}$ & $\mathbf{44.3}$ & $61.0$ & $\mathbf{47.6}$\\
\rowfont{\fontsize{7}{7}\selectfont}
&&& $(10.4)$ & $(9.2)$ & $(19.8)$ & $(3.0)$ & $(8.3)$ & $(1.7)$ & $(2.3)$ & $(9.4)$ & $(6.1)$ & $(10.0)$ & $(3.0)$\\
\cmidrule{3-14} 
\rowfont{\fontsize{9.5}{9.5}\selectfont}
&&\multirow{2}{*}{\textsc{SenSel-InstA}} & $50.4$ & $72.9$ & $40.2$ & $13.9$ & $41.8$ & $44.9$ & $\mathbf{20.9}$ & $72.0$ & $38.3$ & $59.9$ & $45.5$\\
\rowfont{\fontsize{7}{7}\selectfont}
&&& $(6.5)$ & $(5.3)$ & $(17.1)$ & $(1.6)$ & $(7.1)$ & $(0.9)$ & $(3.3)$ & $(4.8)$ & $(5.8)$ & $(5.0)$ & $(1.6)$\\
\cmidrule[0.7pt]{2-14}
\rowfont{\fontsize{9.5}{9.5}\selectfont}
&\multirow{6}{*}{\textsc{No Inst}} &\multirow{2}{*}{\textsc{MaxProb}} & $46.7$ & $80.9$ & $\mathbf{63.7}$ & $33.2$ & $27.5$ & $22.2$ & $22.9$ & $62.9$ & $17.6$ & $56.1$ & $43.4$\\
\rowfont{\fontsize{7}{7}\selectfont}
&&& $(4.6)$ & $(11.4)$ & $(2.6)$ & $(5.1)$ & $(10.4)$ & $(7.7)$ & $(4.2)$ & $(6.5)$ & $(4.0)$ & $(7.6)$ & $(1.4)$\\
\cmidrule{3-14} 
\rowfont{\fontsize{9.5}{9.5}\selectfont}
&&\multirow{2}{*}{\textsc{Entropy}} & $\mathbf{47.2}$ & $80.9$ & $62.3$ & $33.3$ & $\mathbf{27.9}$ & $22.2$ & $22.8$ & $62.9$ & $\mathbf{18.6}$ & $56.1$ & $43.4$\\
\rowfont{\fontsize{7}{7}\selectfont}
&&& $(4.4)$ & $(11.4)$ & $(2.9)$ & $(5.0)$ & $(9.7)$ & $(7.7)$ & $(4.0)$ & $(6.5)$ & $(4.0)$ & $(7.6)$ & $(1.5)$\\
\cmidrule{3-14}
\rowfont{\fontsize{9.5}{9.5}\selectfont}
&&\multirow{2}{*}{\textsc{SenSel-ExOrd}} & $44.4$ & $\mathbf{85.0}$ & $61.9$ & $\mathbf{36.3}$ & $27.6$ & $\mathbf{22.4}$ & $\mathbf{23.3}$ & $\mathbf{68.0}$ & $16.0$ & $\mathbf{57.1}$ & $\mathbf{44.2}$\\
\rowfont{\fontsize{7}{7}\selectfont}
&&& $(6.0)$ & $(9.9)$ & $(2.6)$ & $(4.2)$ & $(11.5)$ & $(8.4)$ & $(5.4)$ & $(12.7)$ & $(4.2)$ & $(12.3)$ & $(2.3)$\\
\bottomrule
\end{tabu}
\caption{\label{tab:auc-ccpc-gptneo} We compare our \textsc{SenSel} method to the \textsc{MaxProb} baseline and the \textsc{Entropy} baseline on the \textsc{GPT-Neo-2.7B} model. \textsc{SenSel} consistently outperforms both baselines under both the \textsc{Inst} setting and the \textsc{No Inst} setting. The standard deviation across five randomly sampled sets of few-shot examples is reported in parenthesis.
}
\end{table*}

\end{document}